\documentclass[12pt]{article}
\usepackage{amsmath}
\usepackage{array}
\usepackage{booktabs}
\usepackage{graphicx}
\usepackage{enumerate}
\usepackage{mathrsfs}
\usepackage{mathtools}
\usepackage{multicol}
\usepackage{multirow}
\usepackage{bm}
\usepackage{amsfonts}

\usepackage[linesnumbered,ruled]{algorithm2e}

\newtheorem{assumption}{Assumption}
\newcommand{\indep}{\perp \!\!\! \perp}
\usepackage{natbib}
\usepackage{url} 
\usepackage[pagewise]{lineno}

\newcommand{\blind}{0}

\begin{document}

\def\spacingset#1{\renewcommand{\baselinestretch}%
{#1}\small\normalsize} \spacingset{1}


\if0\blind
{
  \title{\bf An AI-powered Bayesian generative modeling approach for causal inference in observational studies}
  \author{Qiao Liu\\
    Department of Biostatistics, Yale University\\
    and \\
    Wing Hung Wong \\
    Department of Statistics, Stanford University
    }
  \maketitle
} \fi

\if1\blind
{
  \bigskip
  \bigskip
  \bigskip
  \begin{center}
    {\LARGE\bf An AI-powered Bayesian generative modeling approach for causal inference in observational studies}
\end{center}
  \medskip
} \fi

\bigskip
\begin{abstract}
Causal inference in observational studies with high-dimensional covariates presents significant challenges. We introduce CausalBGM, an AI-powered Bayesian generative modeling approach that captures the causal relationship among covariates, treatment, and outcome. The core innovation is to estimate the individual treatment effect (ITE) by learning the individual-specific distribution of a low-dimensional latent feature set (e.g., latent confounders) that drives changes in both treatment and outcome. This individualized posterior representation yields estimates of the individual treatment effect (ITE) together with well-calibrated posterior intervals while mitigating confounding effect. CausalBGM is fitted through an iterative algorithm to update the model parameters and the latent features until convergence. This framework leverages the power of AI to capture complex dependencies among variables while adhering to the Bayesian principles. Extensive experiments demonstrate that CausalBGM consistently outperforms state-of-the-art methods, particularly in scenarios with high-dimensional covariates and large-scale datasets. By addressing key limitations of existing methods, CausalBGM emerges as a robust and promising framework for advancing causal inference in a wide range of modern applications. The code for CausalBGM is available at \url{https://github.com/liuq-lab/bayesgm}. The document for using CausalBGM is available at \url{https://bayesgm.readthedocs.io}. Supplementary materials for this article are available online.

\end{abstract}

\noindent%
\vfill

\newpage
\spacingset{1.9} 
\section{Introduction}

One central goal for causal inference in observational studies is to estimate the causal effect of one variable (e.g., treatment) on another (e.g., outcome) while accounting for covariates that represent all other measured variables \citep{Pearl2009-PEACII, imbens2015causal,ding2024first}. Covariates are often high-dimensional for modern applications in genomics, economics, and healthcare \citep{prosperi2020causal,davey2020mendel}, which makes the covariate adjustment difficult due to the ``curse of dimensionality'' \citep{d2021overlap}. Additionally, large sample sizes, as is often the case in those scenarios, can further complicate the process by making traditional methods computationally intensive and slow to converge, highlighting the need for developing scalable and effective causal inference method.

To handle the issue of high-dimensional covariates, several dimension reduction methods have been proposed to alleviate the difficulty. For example, one of the most popular approaches is to do adjustment or matching based on the propensity score \citep{rubin1974estimating, rosenbaum_rubin1983, hirano2004propensity}, which is a one-dimensional feature (e.g., a scalar), denoting the probability of receiving a particular treatment given observed covariates. These methods require fitting a propensity score model first, which is typically done by fitting a logistic regression or a machine learning model \citep{lee2010improving}. Another type of dimension reduction method is sufficient dimension reduction (SDR) \citep{li1991sliced,li1992principal}, which projects covariates into a lower-dimensional space, assuming conditional independence of treatment and outcome given the projected features \citep{ghosh2021sufficient,luo2017estimating}. However, SDR-based causal inference methods often restrict dimension reduction to be linear transformations and apply separate projections for each treatment value, limiting its applicability in settings with continuous treatments or complex dependencies. The latent factor approach has also been used as surrogate confounders to adjust for biases in causal effect estimation caused by unobserved confounders \citep{yuan2024confounding}. 

Recently, the rapid development of AI-powered causal inference approaches has shown promising results for causal effect estimation \citep{berrevoets2023causal,lagemann2023deep}. These AI-based approaches typically leverage deep learning techniques and demonstrate superior power in modeling complex dependency and estimation accuracy when the sample size is large. In particular, the Causal \underline{E}ncoding \underline{G}enerative \underline{M}odeling approach, CausalEGM \citep{Liu_pnas2024}, developed by our group, combines anto-encoding and generative modeling to enable nonlinear, structured dimension reduction in causal inference. CausalEGM stands at the intersection of AI and causal inference and has been shown to provide superior performance for developing deep learning-based estimates for the structural equation modeling that describes the causal relations among variables.

\begin{figure}[t]
\centering
\includegraphics[width=0.9\linewidth]{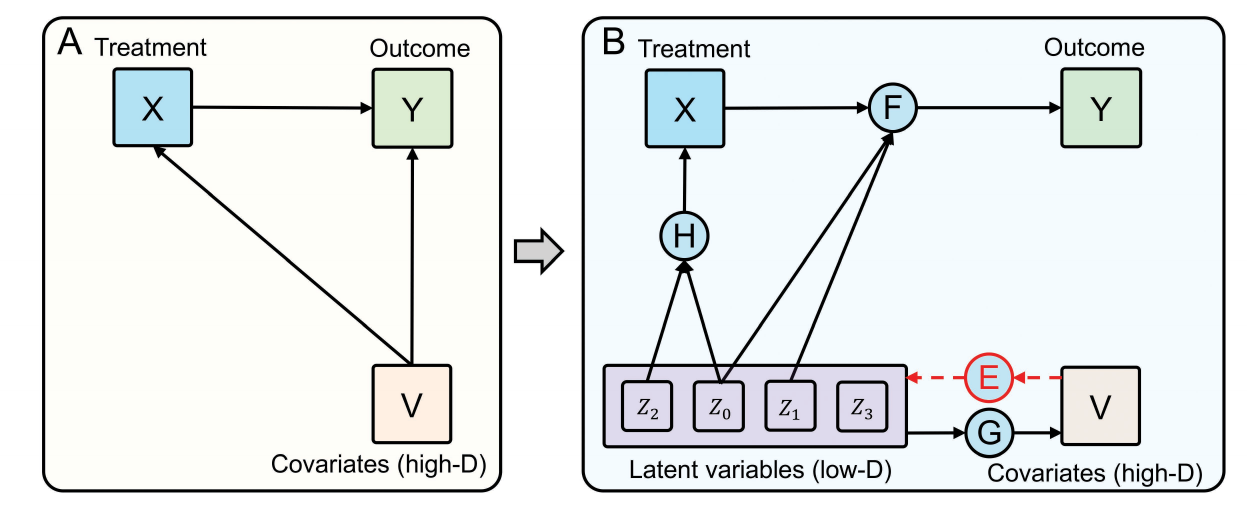}
\caption{Illustration of CausalBGM framework. (A) The typical causal diagram in the observational study where the treatment, outcome, and covariates are observed variables. (B) The overview of CausalBGM model where variables are in rectangles and functions are in circles with incoming arrows indicating inputs to the function and outgoing arrows indicating outputs. $G$, $H$, and $F$ represent generative models for covariates, treatment, and outcome variables, respectively. $E$ represents the encoding function that creates circularity and is used for initialization purpose only. $E$ is removed in CausalBGM during the model training. \label{fig_1}}
\end{figure}

Despite its strong empirical performance, there are two key limitations of the CausalEGM architecture from a Bayesian perspective. First, the joint use of an encoder and a generative decoder introduces a structural loop (dotted arrow in Figure~\ref{fig_1}B). Such circularity violates the acyclicity assumption that is fundamental to Bayesian networks and causal diagrams. Without carefully ensuring a proper directed acyclic graph (DAG) structure, the learned model may struggle to reflect genuine causal relationships. Second, similar to existing AI-based methods primarily focuses on point estimate. CausalEGM relies on deterministic functions to establish the mapping between observed data and latent features. Deterministic mappings can limit the model’s ability to capture and quantify uncertainty, thereby undermining the statistical rigor of the approach and making it challenging to draw reliable causal conclusions in many applications where uncertainty plays a critical role. Probabilistic modeling, in contrast, provides well-defined uncertainty quantification and more robust inference, ensuring that the predictive distributions of the causal effect estimates reflect the true underlying uncertainty in the causal mechanism.

To address the above issues, we introduce CausalBGM, a AI-powered \underline{B}ayesian \underline{G}enerative \underline{M}odeling (BGM) framework for estimating causal effects in the presence of high-dimensional covariates. Compared to CausalEGM, the new CausalBGM removes the encoder function entirely and employs a fully Bayesian procedure to infer latent features (Figure~\ref{fig_1}B without the dotted arrow). By eliminating the encoder-decoder loop, CausalBGM guarantees a clear DAG structure that is consistent with statistical modeling principles. Both the latent variables and model parameters are drawn from probabilistic distributions rather than being deterministically encoded, allowing for the incorporation of prior information and the generation of posterior distributions that more accurately represent uncertainty. By leveraging this fully Bayesian methodology, CausalBGM achieves substantial improvements, providing a principled alignment with Bayesian causal inference \citep{li2023bayesian}. The model eliminates problematic cycles, adopts Bayesian inference, and ultimately provides a more robust and interpretable framework for estimating causal effects in complex, high-dimensional data settings. We highlight several key innovations of CausalBGM as follows.

First, traditional iterative sampling methods (e.g., Gibbs sampling) typically require evaluating conditional distributions that depend on the full dataset at each iteration, which is computationally intensive and often impractical for large-scale datasets. In contrast, CausalBGM introduces a novel iterative algorithm that computes the likelihood only on the current sample or a mini-batch of samples in each iteration step, significantly enhancing scalability. Sampling low-dimensional latent features for each individual is fully decoupled, enabling efficient parallelization and further improving computational efficiency. 

Second, as shown in Section~\ref{sec:init}, iterative sampling of latent features and model parameters can often exhibit suboptimal convergence and performance. To address this, we propose to initiate the updates using estimates from the generative functions obtained by the CausalEGM method, which has strong empirical performance and proven theoretical properties. This strategy ensures a strong starting point for the model, facilitating more stable and accurate iterative updates. Experimental results consistently demonstrate that the EGM initialization significantly enhances predictive accuracy and stability across diverse datasets, underscoring its critical role in achieving superior performance.

Third, instead of directly updating model parameters as deterministic values as standard practice in AI, CausalBGM treats them as random variables and iteratively updates their posterior distributions to account for model uncertainty or variation. Besides, while many existing AI-driven causal inference methods, including CausalEGM, focus solely on modeling the mean function, CausalBGM simultaneously models both the mean and variance functions of observed variables. By incorporating variance modeling, CausalBGM captures a more comprehensive representation of data variability, allowing for the construction of well-calibrated posterior intervals for causal effect estimates.

These innovations uniquely position CausalBGM as a scalable, statistically rigorous, and interpretable framework, bridging the gap between AI and Bayesian causal inference. By addressing key limitations of existing methods, CausalBGM achieves superior performance across a wide range of scenarios, offering a versatile and robust solution for tackling complex causal inference challenges in modern applications.

\section{Methods}
\label{sec:meth}

\subsection{Problem Setup}
Our goal is to estimate the causal effect of one variable $X$ on another variable $Y$ given the presence of the variable $V$ in an observational study based on $i.i.d.$ observations of $\{(X_i,Y_i,\textit{{V}}_i)|i=1,...,N\}$. $X$ denotes the treatment or exposure variable and $Y$ denotes the outcome or response variable. $\textit{{V}}\in \mathbb{R}^p$ represents the covariates in a $p$-dimensional space. $Y\in \mathscr{Y}$ is typically real-valued where the support $\mathscr{Y}$ is a bounded interval in $\mathbb{R}$. $X\in \mathscr{X}$ can be either discrete or continuous where the support $\mathscr{X}$ is either a finite set or a bounded interval in $\mathbb{R}$. 

In order to investigate how the potential outcome will respond to the change of treatment, our primary interest is in estimating the population average of this outcome function, also known as the average dose-response function (ADRF), defined by: 
\begin{equation} 
\label{eqn_mu}
\mu(x)=\mathbb{E}[Y(x)].
\end{equation}
Since we only observe the potential outcomes indexed by the treatment variable (e.g., factual outcome). The random variable $Y(x)$ is not directly observable due to the counterfactual outcomes, and the expectation $\mu(x)$ cannot generally be directly identified from the joint distribution of the observed data  $(X,Y,\textit{{V}})$. Therefore, additional assumptions are required to ensure the identifiability of $\mu(x)$.

We first assume $X$, $Y$, and $V$ are generated by a latent variable $Z \in \mathbb{R}^q$ where $q\ll p$. We denote $Z_0$ as a subset of the latent variable $Z$, which affects both treatment and outcome. Next, we introduce a modified version of the ``unconfoundedness" assumption with respect to the latent confounding variable $Z_0$.

\begin{assumption}\label{as:modified unconfounderness}   
\textbf{(Unconfoundedness)} Given the low-dimensional latent confounding variable $Z_0$, the potential outcomes $Y(x)$ is independent of treatment variable $X$,

\begin{equation} 
\label{eqn:3}
X\indep Y(x)|Z_0.
\end{equation}
\end{assumption}
Under the traditional "unconfoundedness" assumption, one typically conditions on the high-dimensional covariates $V$. However, our Assumption \ref{as:modified unconfounderness} makes this requirement less restrictive by showing that it is sufficient to condition on a low-dimensional feature set representing the covariates. Once $Z_0$ is given, there should be no unobserved confounding variables that induce correlated changes between the treatment and the outcome. 

Based on assumption \ref{as:modified unconfounderness}, it follows that the ADRF can be identified through the following equation, 
\begin{equation} \label{eqn_adrf}
\mu(x)=\int \mathbb{E}[Y|X=x,Z_0=z_0]p_{Z_0}(z_0)dz_0.
\end{equation}
The identification proof is given in Appendix A. Equation \ref{eqn_adrf} transforms the causal inference problem into the problem of learning a latent confounding variable $Z_0$ given the observational data. In the following section, we will outline a AI-powered Bayesian framework in order to learn $Z_0$ and estimate the $\mu(x)$ in equation \ref{eqn_adrf}.

\subsection{Causal Generative Modeling} 
Our model is described in Figure~\ref{fig_1}, where $X,Y,V$ represents observed variables and $Z=(Z_0,Z_1,Z_2,Z_3)$ denotes the low-dimensional latent variable that needs to be learned. The whole latent space is partitioned into four parts that play different roles in the following generative models of $X$, $Y$, and $V$.
\begin{equation}
\label{eqn:generative_model}
\left\{
\begin{aligned}
Z \sim& \pi_Z(Z),\\
\theta_X\sim& \pi_{\theta_X}(\theta_X),\theta_Y\sim \pi_{\theta_Y}(\theta_Y),\theta_V\sim \pi_{\theta_V}(\theta_V),\\
V \sim& P(V|Z;\theta_V),\\
X \sim& P(X|Z_0,Z_2;\theta_X),\\
Y \sim& P(Y|X,Z_0,Z_1;\theta_Y),\\
\end{aligned}
\right.
\end{equation}
where $Z_0$ denotes the latent confounding variable that affects both treatment and outcome, $Z_1$ represents the latent features that affect only the outcome, $Z_2$ relates to the latent features that affect only the treatment, and $Z_3$ comprises the remaining latent features that affect neither treatment nor outcome. By partitioning the latent features $Z$ into four different independent components, CausalBGM is able to isolate the underlying dependencies of covariates on treatment and outcome in the low-dimensional latent space. Through the above partition, we aim to identify a minimal covariate feature set (e.g., $Z_0$) that affects both treatment and outcome. $\theta_X$, $\theta_Y$, and $\theta_V$ are the parameters of the three generative models of treatment, outcome, and covariates, respectively. All the prior distributions are set to be standard multivariate normal distributions. 

The three generative models can be flexibly parameterized by any parametric family, such as the exponential family (see Appendix B). In default, we model the conditional distribution as normal distributions for continuous variables and logistic regression for discrete variables. In typical causal inference settings, the generative processes are defined as follows: 

\begin{itemize}
\item \textbf{Covariate Modeling}. The covariate variable $V$ is modeled as a multivariate normal distribution as follows.
\begin{equation}
\label{eqn:gen_cov}
P(V|Z;\theta_V) = \mathcal{N}(\mu_v(Z),\Sigma_v(Z)),
\end{equation}
where both mean and covariance matrix are learnable functions of latent variable $Z$ parameterized by $\theta_V$. To simplify, the covariance matrix $\Sigma_v(Z)$ is represented as $\sigma^{2}_v(Z)I_p$ where $I_p$ is the $p$-dimensional identity matrix and $\sigma^{2}_v(Z)$ is a learnable variance function. 

\item \textbf{Treatment Modeling}. For continuous treatments, the treatment variable $X$ is modeled as:
\begin{equation}
\label{eqn:gen_treat_con}
P(X|Z_0,Z_2;\theta_X) = \mathcal{N}(\mu_x(Z_0,Z_2),\sigma^{2}_x(Z_0,Z_2)),
\end{equation}
where both mean $\mu_x(Z_0,Z_2)$ and variance $\sigma^{2}_x(Z_0,Z_2)$ are learnable functions of $Z_0$ and $Z_2$ parameterized by $\theta_X$. 

For binary treatments, $X$ is modeled using a generalized logistic regression:
\begin{equation}
\label{eqn:gen_treat_dis}
P(X=1|Z_0,Z_2;\theta_X) = 1/(1+e^{-\xi}),
\end{equation}
where $\xi \sim \mathcal{N}(\mu_x(Z_0,Z_2),\sigma^{2}_x(Z_0,Z_2))$, and the resulting probability is equivalent to the propensity score.

\item \textbf{Outcome Modeling}. The outcome variable $Y$ is modeled as a normal distribution:
\begin{equation}
\label{eqn:gen_outcome}
P(Y|X,Z_0,Z_1;\theta_Y) = \mathcal{N}(\mu_y(X,Z_0,Z_1),\sigma^{2}_y(X,Z_0,Z_1)),
\end{equation}
where both mean $\mu_y(X,Z_0,Z_1)$ and variance $\sigma^{2}_y(X,Z_0,Z_1)$ are learnable functions of $X$, $Z_0$ and $Z_1$, parameterized by $\theta_Y$.

\end{itemize}

Note that the learnable functions ($\mu_x, \sigma^2_x$),($\mu_y, \sigma^2_y$), and ($\mu_v, \sigma^2_v$) are represented by three Bayesian neural networks~\citep{jospin2022hands}, parameterized by $\theta_X$, $\theta_Y$, and $\theta_V$ respectively. In the next section, we will illustrate how we learn the distribution of model parameters $\theta_X$, $\theta_Y$, $\theta_V$ in order to account for the model uncertainty or variation.

\subsection{Iterative Updating Algorithm} 

We adopted a stochastic iterative algorithm \citep{liu2026bayesian} to update the posterior distribution of model parameters and the posterior distribution of latent variable $Z$ until convergence. According to Bayes' theorem, the joint posterior distribution of the latent features and model parameters is represented as
\begin{equation}\label{eqn:posterior}
\begin{aligned}
&P(Z,\theta_X,\theta_Y,\theta_V|X,Y,V)=P(\theta_X,\theta_Y,\theta_V|X,Y,V)P(Z|X,Y,V,\theta_X,\theta_Y,\theta_V). \\
\end{aligned}
\end{equation}
Since the true joint posterior is intractable, we approximate the problem by designing an iterative algorithm. Specifically, we iteratively 1) update the posterior distribution of latent variable $Z$ from $P(Z|X,Y,V,\theta_X,\theta_Y,\theta_V)$. 2) update the posterior distribution of model parameters $(\theta_X,\theta_Y,\theta_V)$ from $P(\theta_X,\theta_Y,\theta_V|X,Y,V,Z)$. 

To estimate the posterior distribution of the latent variable $Z$ in step 1), we denote the log-posterior of the latent variable $Z$ as 
\begin{equation}\label{eqn:posterior_Z}
\begin{aligned}
&logP(Z|X,Y,V,\theta_X,\theta_Y,\theta_V)=log\pi_Z(Z)+logP(X,Y,V|Z,\theta_X,\theta_Y,\theta_V)+C, \\
&=log\pi_Z(Z)+logP(V|Z,\theta_X,\theta_Y,\theta_V)+logP(X,Y|Z,\theta_X,\theta_Y,\theta_V)+C,  \\
&= log\pi_Z(Z)+logP(V|Z;\theta_V)+logP(X|Z_0,Z_2;\theta_X)+logP(Y|X,Z_0,Z_1;\theta_Y)+C,   \\ 
\end{aligned}
\end{equation}
where $C=log\pi_{\theta_X}(\theta_X)+log\pi_{\theta_Y}(\theta_Y)+log\pi_{\theta_V}(\theta_V)-logP(X,Y,V,\theta_X,\theta_Y,\theta_V)$ is irrelevant to $Z$. The second equality in (\ref{eqn:posterior_Z}) is obtained by the conditional independence in Assumption \ref{as:modified unconfounderness}. The log-likelihood of the three generative models are denoted as 
\begin{equation}
\label{eqn:ll_models}
\left\{
\begin{aligned}
&logP(V|Z;\theta_V)=-\frac{p}{2}log(\sigma^2_v(Z))-\frac{1}{2\sigma^2_v(Z)}||V-\mu_v(Z)||_2^2+C_1,\\
&logP(X|Z_0,Z_2;\theta_X)=-\frac{1}{2}log(\sigma^2_x(Z_0,Z_2))-\frac{1}{2\sigma^2_x(Z_0,Z_2)}(X-\mu_x(Z_0,Z_2))^2+C_2,\\ 
&logP(Y|X,Z_0,Z_1;\theta_Y)=-\frac{1}{2}log(\sigma^2_y(X,Z_0,Z_1))-\frac{1}{2\sigma^2_y(X,Z_0,Z_1)}(Y-\mu_y(X,Z_0,Z_1))^2+C_3,
\end{aligned}
\right.
\end{equation}
where $C_1$,$C_2$, and $C_3$ are constants. 

To update the posterior $P(\theta_X,\theta_Y,\theta_V|X,Y,V,Z)$ over all model parameters $\theta_X$, $\theta_Y$, and $\theta_V$ from three generative models in step 2), we further decompose the joint posterior for the model parameters based on conditional independence, which is denoted as 
\begin{equation}
\label{eqn:posterior_theta}
\left\{
\begin{aligned}
&logP(\theta_X|X,Y,V,Z)=log\pi_{\theta_X}(\theta_X)+logP(X|Z_0,Z_2;\theta_X)+C_4,\\
&logP(\theta_Y|X,Y,V,Z)=log\pi_{\theta_Y}(\theta_Y)+logP(Y|X,Z_0,Z_1;\theta_Y)+C_5,\\ 
&logP(\theta_V|X,Y,V,Z)=log\pi_{\theta_V}(\theta_V)+logP(V|Z;\theta_V)+C_6,\\
\end{aligned}
\right.
\end{equation}
where $C_4$ is irrelevant with $\theta_X$, $C_5$ is irrelevant with $\theta_Y$, and $C_6$ is irrelevant with $\theta_V$. Since the posterior distribution of parameters in each generative model is intractable, we employ three Bayesian network networks, which use variational inference (VI) to approximate each term in (\ref{eqn:posterior_theta}). Specifically, we introduce three variational distributions $q_{\phi_X}(\theta_X)$,$q_{\phi_Y}(\theta_Y)$, and $q_{\phi_V}(\theta_V)$ to approximate the true posteriors in (\ref{eqn:posterior_theta}), respectively. The variational distributions are chosen to be normal distributions as $q_{\phi_X}(\theta_X) \sim \mathcal{N}(\theta_X|\mu_{\phi_X}, \sigma_{\phi_X}^2)$, $q_{\phi_Y}(\theta_Y) \sim \mathcal{N}(\theta_Y|\mu_{\phi_Y}, \sigma_{\phi_Y}^2)$, and $q_{\phi_V}(\theta_V) \sim \mathcal{N}(\theta_V|\mu_{\phi_V}, \sigma_{\phi_V}^2I_p)$. Note that $\phi_X=(\mu_{\phi_X},\sigma_{\phi_X}^2)$, $\phi_Y=(\mu_{\phi_Y},\sigma_{\phi_Y}^2)$, and $\phi_V=(\mu_{\phi_V},\sigma_{\phi_V}^2)$ are learnable parameters for the variational distributions (variational parameters). The evidence lower bound (ELBO) for each posterior is defined as 
\begin{equation}
\label{eqn:elbo}
\left\{
\begin{aligned}
&\mathcal{L}(\phi_X)=\mathbb{E}_{q_{\phi_X}(\theta_X)}[logP(X|Z_0,Z_2;\theta_X)]-KL(q_{\phi_X}(\theta_X)||\pi_{\theta_X}(\theta_X)),\\
&\mathcal{L}(\phi_Y)=\mathbb{E}_{q_{\phi_Y}(\theta_Y)}[logP(Y|X,Z_0,Z_1;\theta_Y)]-KL(q_{\phi_Y}(\theta_Y)||\pi_{\theta_Y}(\theta_Y)),\\ 
&\mathcal{L}(\phi_V)=\mathbb{E}_{q_{\phi_V}(\theta_V)}[logP(V|Z;\theta_V)]-KL(q_{\phi_V}(\theta_V)||\pi_{\theta_V}(\theta_V)),\\
\end{aligned}
\right.
\end{equation}
where the first term in the ELBO denotes the expected log-likelihood under the variational distribution and the second term denotes Kullback-Leibler divergence between the variational posterior and the prior distribution over model parameters. To facilitate the computation of gradient w.r.t the variational parameters, we use reparameterization trick, which is represented as
\begin{equation}
\label{eqn:flipout}
\left\{
\begin{aligned}
&\hat{\theta}_X=\mu_{\phi_X}+\sigma_{\phi_X}\odot\epsilon_X,\\
&\hat{\theta}_Y=\mu_{\phi_Y}+\sigma_{\phi_Y}\odot\epsilon_Y,\\ 
&\hat{\theta}_V=\mu_{\phi_V}+\sigma_{\phi_V}\odot\epsilon_V,\\
\end{aligned}
\right.
\end{equation}
where $\epsilon_X \sim \mathcal{N}(0,I_{d_X})$, $\epsilon_Y \sim \mathcal{N}(0,I_{d_Y})$, and $\epsilon_V \sim \mathcal{N}(0,I_{d_V})$. $\odot$ is the element-wise product. $d_X$, $d_Y$, and $d_V$ denote the number of parameters in each generative model. Using variational inference in BNNs for each mini-batch may lead to high-variance gradient estimates. We adopt the Flipout technique \citep{wen2018flipout} in the implementation of the reparameterization trick to reduce this variance by decorrelating the model parameters perturbations across different training examples in the same mini-batch. Briefly, in stead of using a single shared random draw of model parameters for the entire mini-batch, Flipout constructs pseudo-independent perturbations for each example independently within a mini-batch, which decorrelates the gradients, reduces the variance, and stabilizes the training process. 

Note that we update the posterior distribution of model parameters for treatment model, covariate model, and outcome model sequentially. For each generative model, we first update the variational parameters to maximize the ELBO in (\ref{eqn:elbo}) and then sample model parameters from (\ref{eqn:flipout}). Given the sampled model parameters, the regular forward pass through the network (e.g., each layer contains matrix multiplication followed by non-linear activation function) is computed to get the mean and variance parameters in (\ref{eqn:gen_cov}-\ref{eqn:gen_outcome}).

Each iteration only requires the current sample or a random mini-batch of observed samples. During the iteration algorithm, we first take a derivative of equation (\ref{eqn:posterior_Z}) w.r.t the latent variable $Z$ and employ a stochastic gradient descent (SGD) to update the latent variable $Z$ for each individual given the current model parameters. Then, we take a derivative of each ELBO term in (\ref{eqn:elbo}) w.r.t the variational parameters ($\phi_X$, $\phi_Y$, or $\phi_V$) in the three generative models sequentially and employ a stochastic gradient ascent to update the variational parameters given the current latent variables to maximize the ELBO. During test stage, we only need to infer the posterior distribution of latent variable $Z$ given the test data. To achieve this, we first sampled model parameters ($\theta_X$, $\theta_Y$, and $\theta_V$) from the variational distribution parameterized by $\phi_X$, $\phi_Y$, and $\phi_V$ through (\ref{eqn:flipout}). Then we use Markov chain Monte Carlo (MCMC) method \citep{liu2001monte} to sample from the posterior distribution in (\ref{eqn:posterior_Z}) for each individual. We choose the standard Metropolis–Hastings algorithm \citep{robert2004metropolis} as default. Note that this individual-level sampling process is fully decoupled, enabling parallelization and improving computational efficiency. The causal effect and the corresponding posterior interval with user-specified
significant level, can be then estimated based on the MCMC samples of latent variable and the learned generative models. 

In the binary treatment setting, the individual treatment effect (ITE) for the $i^{th}$ unit is estimated as
\begin{equation} \label{eqn:ite}
\hat{\Delta}_i=\frac{1}{S}\sum_{s=1}^S(\hat{y}_{i,s}^{(1)}-\hat{y}_{i,s}^{(0)}),
\end{equation}
where $\hat{y}_{i,s}^{(1)}\sim \mathcal{N}(\mu_y(X=1,Z_0=z_{0,i}^s,Z_1=z_{1,i}^s),\sigma^2_y(X=1,Z_0=z_{0,i}^s,Z_1=z_{1,i}^s))$ and $\hat{y}_{i,s}^{(0)}\sim \mathcal{N}(\mu_y(X=0,Z_0=z_{0,i}^s,Z_1=z_{1,i}^s),\sigma^2_y(X=0,Z_0=z_{0,i}^s,Z_1=z_{1,i}^s))$. Note that $\{z_{i}^{s}=(z_{0,i}^{s},z_{1,i}^{s},z_{2,i}^{s},z_{3,i}^{s})\}_{s=1}^{S}$ denotes the MCMC samples of the latent variable $Z$ from the $i^{th}$ unit. Equation (\ref{eqn:ite}) represents an unbiased estimation of ITE using the MCMC samples. The posterior interval for ITE can then be constructed. Given a desired significant level $\alpha$ (e.g., $\alpha=0.05$), we calculate the quantile to represent the lower and upper posterior interval bounds that meet the desired significant level as 
\begin{equation}
\label{eqn:quantile_ite}
\left\{
\begin{aligned}
\hat{L}_{\Delta_i}=&Quantile_{\alpha/2}(\{(\hat{y}_{i,s}^{(1)}-\hat{y}_{i,s}^{(0)})\}_{s=1}^S),\\
\hat{U}_{\Delta_i}=&Quantile_{1-\alpha/2}(\{(\hat{y}_{i,s}^{(1)}-\hat{y}_{i,s}^{(0)})\}_{s=1}^S).\\ 
\end{aligned}
\right.
\end{equation}
where $Quantile_{\alpha/2}(\cdot)$ is the quantile function of the sampling distribution that cuts off the lower $\alpha/2$ tail of the distribution.

In the continuous treatment setting, the ADRF is estimated by 
\begin{equation} \label{eqn:adrf}
\hat{\mu}(x)=\frac{1}{S\times N}\sum_{s=1}^{S}\sum_{i=1}^{N}\hat{y}_{i,s}(x),
\end{equation}
where $\hat{y}_{i,s}(x)\sim \mathcal{N}(\mu_y(X=x,Z_0=z_{0,i}^s,Z_1=z_{1,i}^s),\sigma^2_y(X=x,Z_0=z_{0,i}^s,Z_1=z_{1,i}^s))$. Equation (\ref{eqn:adrf}) represents an unbiased estimation of ADRF using all the MCMC samples. Similarily, the lower and upper posterior interval bounds of $\hat{\mu(x)}$ that satisfy a desired significant level $\alpha$ are estimated by

\begin{equation}
\label{eqn:quantile_adrf}
\left\{
\begin{aligned}
\hat{L}_{\mu}(x)=&Quantile_{\alpha/2}(\{\frac{1}{ N}\sum_{i=1}^{N}\hat{y}_{i,s}(x)\}_{s=1}^S),\\
\hat{U}_{\mu}(x)=&Quantile_{1-\alpha/2}(\{\frac{1}{ N}\sum_{i=1}^{N}\hat{y}_{i,s}(x)\}_{s=1}^S).\\ 
\end{aligned}
\right.
\end{equation}

\subsection{Choice of latent dimension}
The latent space is partitioned into four independent parts that play different roles in the three generative models for treatment, covariates, and outcome variables. The previous work CausalEGM has demonstrated the robustness of such partition with respect to variations in the dimensionality of latent features. Here, we provide an intuitive strategy based on sufficient dimension reduction (SDR) to help determine the dimensionality of latent features. SDR aims to identify a $k$-dimensional subspace of the $p$-dimensional predictors ($k\ll p$) that captures all the information about a scalar response. Here, we use sliced inverse regression (SIR) \citep{li1991sliced} that employs the covariance structure of the conditional expectations of predictors given response. We compute eigenvalues of the estimated covariance matrix from SIR and retain components by inspecting eigenvalue decay and cumulative variance. Considering that linear methods such as SIR may underestimate $k$ as they fail to capture nonlinear dependencies effectively. Here, we use a conservative strategy by using SIR $\mathbb{E}[V|X]$ to estimate $dim(Z_2)$ and using SIR $\mathbb{E}[V|Y]$ to estimate $dim(Z_1)$. A similar eigenvalue analysis for the covariance matrix of $V$ is conducted to estimate the total dimension of the latent space $dim(Z)$. The dimension of latent confounder $dim(Z_0)$ is a model hyperparameter.

\subsection{Model Initialization}

The parameters of neural networks (e.g., weights and biases) are typically initialized by a uniform or normal distribution. However, as shown by our experiments (Table~\ref{table:init}), the model performance can be further improved in most cases through our designed innovative strategy for model parameters initialization, inspired from the encoding generative modeling (EGM) \citep{Liu_pnas2024}, compared to the traditional neural network initialization. An additional encoder function $E$, represented by a Bayesian neural network, is added to CausalBGM to directly map the covariate $V$ to the latent variable (dotted line in Figure~\ref{fig_1}). Specifically, we desire that the distribution of $Z=E(V)$ should match a pre-specified distribution, which is set to be a standard normal distribution (e.g., prior of $Z$). The distribution match is achieved by adversarial training \citep{goodfellow2014generative,liu2021density}. By the encoding process, the high-dimensional covariates with unknown distribution are mapped to a low-dimensional latent space with a desired distribution. Since the generative models in CausalBGM include both the mean and variance functions. We add additional The $L_2$ regularization of all the variance terms to ensure reasonably small variance in each generative model during the initialization process.

\subsection{Model Hyperparameters}
CausalBGM contains three generative models, which are represented by three Bayesian neural networks (BNNs), respectively. The BNN for covariate $V$ contains 5 hidden Bayesian layers and each layer has 64 hidden nodes. The output of BNN for covariate $V$ is $(p+1)$-dimensional where the first $p$ digits denote the mean $\mu_v$ and the last digit denotes variance $\sigma^2_v$. The BNN for treatment $X$ and outcome $Y$ contains 3 hidden Bayesian layers with 64, 32, and 8 hidden nodes. The output of BNN of treatment $X$ and outcome $Y$ is 2 dimensional, representing the mean and variance, respectively. The leaky-ReLu function ($LeakyReLU(x)=max(0.2x,x)$) is used as the non-linear activation function in each hidden layer. The Softplus non-linear activation function ($Softplus(x)=log(1+e^x)$) is applied to the last digit of the BNN output to ensure positivity of variance. Adam optimizers \citep{kingma2014adam} with learning rate $0.0001$ are used to update latent variable and model parameters, respectively. The CausalBGM model is trained in a mini-batch manner with batch size 32. The default training epochs of CausalBGM with random initilization strategy is 500. If EGM initilization strategy is used, we initialize model parameters of CausalBGM by conducting EGM for $30,000$ mini-batches as default. After model initialization, the encoder $E$ as a ``shortcut" to learn the latent variable is removed during the follow-up CausalBGM training with an iterative approach for up to 100 epochs. In the random walk Metropolis algorithm, we set the proposal distribution to be a normal distribution centered at the current sample with covariance matrix $I_q$. The Markov chain samples from the first $5,000$ iterations are discarded so that the effect of initial values is minimized (burn-in stage). Then we run the Markov chains parallelly for all samples until $3,000$ MCMC latent samples are collected for each sample.

\section{Results}
\label{sec:res}

We conducted a range of experiments to evaluate the performance of CausalBGM against various state-of-the-art methods across both continuous and binary treatment scenarios. In the continuous treatment setting, our focus was on assessing how well CausalBGM could learn the average dose–response function (ADRF) that describes the change of outcome variable in response to the treatment or exposure variable. In the binary treatment setting, we aimed to verify CausalBGM’s ability to estimate both the population-level average treatment effect (ATE) and the individual-level treatment effect (ITE).

\subsection{Datasets}

For the continuous treatment setting, we examined four public datasets used in previous studies \citep{hirano2004propensity, sun2015causal, colangelo2020double}, comprising three simulated datasets and one semi-synthetic dataset. Each of the simulation datasets has $20,000$ as the sample size and $200$ covariate features. We focus on the ADRF estimate in a bounded interval. The semi-synthetic data were derived from a sample of 71,345 twin births, where weight served as the continuous treatment variable and the risk of death is treated as the outcome variable, which is simulated from a risk model. Each individual has 50 covariates. In general, the simulation risk model suggests that a higher weight of an infant leads to a lower death rate.

In the binary treatment setting, we employed datasets from the 2018 Atlantic Causal Inference Conference (ACIC) competition, which were constructed from linked birth and infant death records (LBIDD) with 117 measured covariates. These semi-synthetic datasets have treatments and outcomes simulated from diverse data-generating processes. We chose nine datasets that utilized the most complex generation processes (e.g., the highest degree of generation function) with sample sizes spanning from 1,000 to 50,000 observations. Complete details on all datasets can be found in Appendix C.

\subsection{Model Evaluation}

In the continuous treatment setting, the goal is to evaluate whether the ADRF under a bounded interval is accurately estimated. To quantitatively measure the difference between the true ADRF curve and the estimated ADRF curve, two metrics, including root mean squared error (RMSE) and mean absolute percentage error (MAPE), are used for evaluation purposes denoted as 
\begin{equation}
\label{model_evaluation_cont}
\left\{
\begin{aligned}
RMSE=&\sqrt{\frac{1}{K}\sum_{k=1}^K(\mu(x_k)-\hat{\mu}(x_k))^2}, \\
MAPE=&\frac{1}{K}\sum_{k=1}^K|\frac{\mu(x_k)-\hat{\mu}(x_k)}{\mu(x_k)}|. \\
\end{aligned}
\right.
\end{equation}
where $K$ represents the number of different treatment values equally distributed in the bounded interval.

In the binary treatment setting, we aim to evaluate whether individual treatment effect (ITE) can be accurately estimated. We adopt two evaluation metrics, including absolute error of \underline{a}verage \underline{t}reatment \underline{e}ffect ($\epsilon_{ATE}$) and mean squared error of \underline{p}recision in \underline{e}stimation of \underline{h}eterogeneous \underline{e}ffect ($\epsilon_{PEHE}$), which are denoted as 
\begin{equation}
\label{model_evaluation_binary}
\left\{
\begin{aligned}
\epsilon_{ATE} =&|\frac{1}{N}\sum_{i=1}^{N}{}{\hat{\Delta}_i}-\frac{1}{N}\sum_{i=1}^{N}{}{{\Delta}_i}|,\\
\epsilon_{PEHE}=&\frac{1}{N}\sum_{i=1}^{N}(\hat{\Delta}_i-{\Delta}_i)^2, \\
\end{aligned}
\right.
\end{equation}
where $N$ is the sample size, $\hat{\Delta}_i$ and ${\Delta}_i$ denote the estimated and true individual treatment effect (ITE) for the $i^{th}$ unit, respectively.

\subsection{Baseline Methods}

For the continuous treatment setting, we considered four well-established baseline methods: ordinary least squares (OLS), the regression prediction estimator (REG) \citep{schafer2015causal, galagate2016causal, imai2004causal}, double debiased machine learning estimators (DML) \citep{colangelo2020double}, and CausalEGM \citep{Liu_pnas2024}. Note that different machine learning methods shall be used in the DML method. For the binary treatment setting, we compared CausalBGM against seven leading methods for estimating treatment effect, including two variants of CFR \citep{shalit2017estimating}, Dragonnet \citep{shi2019adapting}, CEVAE \citep{louizos2017causal}, GANITE \citep{yoon2018ganite}, CausalForest \citep{wager2018estimation}, and CausalEGM \citep{Liu_pnas2024}. Additional details about these competing methods are provided in Appendix D.

\subsection{Continuous Treatment Experiments}

We conducted comprehensive experiments to evaluate the performance of CausalBGM against a suite of state-of-the-art baseline methods, including the previous CausalEGM framework under continuous treatment settings. The treatment variable $X$ is defined over a bounded interval in $\mathbb{R}$. We simulated three datasets from the previous works with a sample size of 20,000 and 200 covariates. We used the same latent dimensions as those tested for CausalEGM to ensure a fair comparison in all datasets. Specifically, for four distinct data-generating processes, the latent dimensions of ($Z_0$, $Z_1$, $Z_2$, $Z_3$) were set to (1,1,1,7), (2,2,2,4), (5,5,5,5), and (1,1,1,7), respectively.

\begin{figure}[htbp]
  \centering
  \includegraphics[width=0.99\columnwidth]{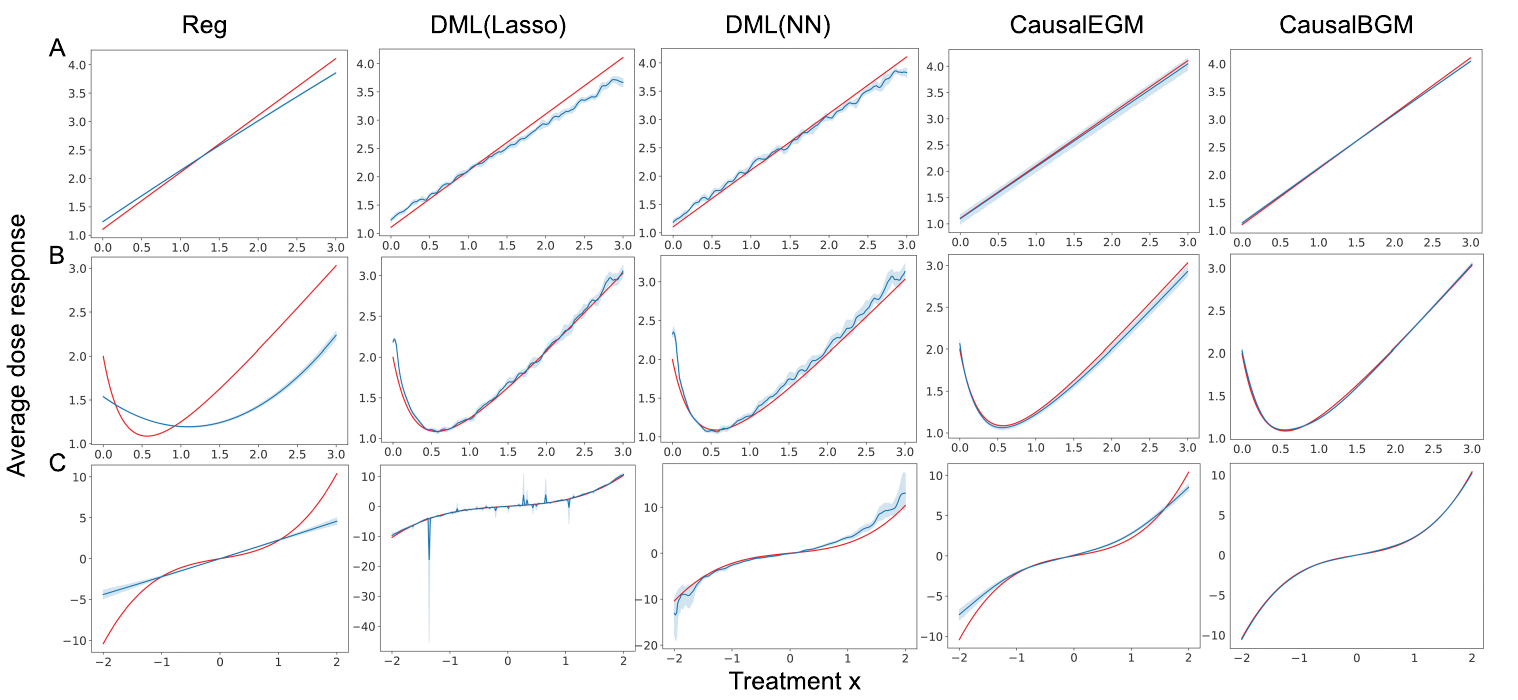}
  \caption{The performance of CausalBGM and baseline methods (Reg, DML with Lasso or neural network, and CausalEGM) under continuous treatment settings across three benchmark datasets. (A) Sun et al. dataset. (B) Hirano and Imbens dataset. (C) Colangelo and Lee dataset. The red curves represent the ground truth, while the blue curves indicate the estimated average dose-response of different methods with 95\% confidence intervals based on 10 independent runs.}
  \label{fig:continuous_result}
\end{figure}

Under these settings, CausalBGM demonstrated superior performance compared to all competing methods, including CausalEGM, REG, and double debiased machine learning estimators using lasso and neural networks, achieving consistently higher accuracy. In comparison to CausalEGM, which already showed significant gains over traditional approaches, CausalBGM further improved the accuracy of the ADRF estimate and reduced both bias and variance by a large margin. As illustrated in Figure~\ref{fig:continuous_result}, the REG continued to produce larger estimation errors and exhibited limited flexibility while the double debiased machine learning estimators displayed undesirable spikes and fluctuations in their dose–response curves. CausalBGM, by contrast, yielded smoother and more stable dose–response estimates, capturing the underlying causal structure more faithfully and with smaller variance. 

Specifically, all methods closely follow the ground truth with linear relationship of Sun et al. dataset, but CausalBGM exhibits the most stable and precise estimations (Figure~\ref{fig:continuous_result}A). In the other two datasets with a non-linear relationship, CausalBGM consistently provides more accurate estimations, particularly at the boundaries of the treatment intervals, where other computing methods display substantial deviations (Figure~\ref{fig:continuous_result}B-C). These results highlight the robustness and accuracy of CausalBGM in capturing complex dose-response relationships, especially in challenging scenarios with non-linear effects.

We further use quantitative metrics to evaluate the performance of different methods across the above three simulation datasets and a semi-synthetic dataset. As shown in Table~\ref{table:continuous}, CausalBGM demonstrates consistently superior performance in estimating average dose-response functions, achieving the state-of-the-art RMSE and MAPE in all cases. For example, CausalBGM reduced the RMSE by half, from 0.074 to 0.037, compared to CausalEGM in Sun et al. dataset. CausalBGM achieved a nearly three-fold improvement in the metric MAPE, reducing the value from 0.035 to 0.013 within the same dataset. These improvements underscore the effectiveness of CausalBGM and ensure more robust and reliable causal inference across a range of complex, high-dimensional continuous treatment settings.

\renewcommand\arraystretch{0.3}
\begin{table}[t!]
  \centering
  \caption{Results for the continuous treatment setting. Each method was run 10 times, and the standard deviation is shown. The best performance is highlighted in bold.}
  \begin{tabular}{ccccl}
    \toprule
    Dataset & Method & RMSE & MAPE \\
    \midrule
    \multirow{7}{*}{Imbens et al.}  
    & OLS & $0.680\pm0.0$ & $0.367\pm0.0$ \\
    & REG & $0.525\pm0.0$ & $0.214\pm0.0$ \\
    & DML(Lasso) & $0.090\pm0.0$ & $0.037\pm0.0$ \\
    & DML(NN) & $0.133\pm0.022$ & $0.052\pm0.011$ \\   
    & CausalEGM & $0.041\pm0.014$ & $0.019\pm0.006$ \\  
    & \textbf{CausalBGM} & $\bm{0.028\pm0.007}$ & $\bm{0.013\pm0.003}$ \\
    \midrule
    \multirow{7}{*}{Sun et al.}     
    & OLS & $0.140\pm0.0$ & $0.041\pm0.0$ \\
    & REG & $0.117\pm0.0$ & $0.039\pm0.0$ \\
    & DML(Lasso) & $0.163\pm0.0$ & $0.050\pm0.0$ \\
    & DML(NN) & $0.097\pm0.019$ & $0.035\pm0.006$ \\
    & CausalEGM & $0.074\pm0.040$ & $0.035\pm0.017$ \\
    & \textbf{CausalBGM} & $\bm{0.037\pm0.009}$ & $\bm{0.013\pm0.005}$ \\
    \midrule
    \multirow{7}{*}{Lee et al.} 
    & OLS & $1.3\pm0.0$ & $1.2\pm0.0$ \\
    & REG & $1.5\pm0.0$ & $0.565\pm0.0$ \\
    & DML(Lasso) & $0.487\pm0.0$ & $0.168\pm0.0$ \\
    & DML(NN) & $1.3\pm0.581$ & $0.494\pm0.181$ \\
    & CausalEGM & $0.125\pm0.040$ & $0.119\pm0.080$ \\
    & \textbf{CausalBGM} & $\bm{0.080\pm0.030}$ & $\bm{0.072\pm0.035}$ \\
    \midrule
    \multirow{7}{*}{Twins} 
    & OLS & $0.109\pm0.0$ & $0.260\pm0.0$ \\
    & REG & $11\pm0.0$ & $64\pm0.0$ \\
    & DML(Lasso) & $0.075\pm0.0$ & $0.165\pm0.0$ \\
    & DML(NN) & $0.059\pm0.002$ & $0.158\pm0.006$ \\
    & CausalEGM & $0.034\pm0.020$ & $0.090\pm0.053$ \\
    & \textbf{CausalBGM} & $\bm{0.031\pm0.007}$ & $\bm{0.077\pm0.009}$ \\
    \bottomrule
  \end{tabular}
\label{table:continuous}
\end{table}
\normalsize

\subsection{Binary Treatment Experiments}

Most causal inference methods target binary treatment settings, which are prevalent in many real-world applications and the treatment variable only takes binary value from $\{0,1\}$. In such a setting, we evaluated CausalBGM alongside several state-of-the-art methods, including TARNET, CFRNET, CEAVE, GANITE, Dragonnet, CausalForest, and CausalEGM across datasets of varying sizes from the ACIC 2018 benchmark. The dimension of latent space is set to be $(3,6,3,6)$, which is the same as CausalEGM. Two evaluation metrics, including $\epsilon_{ATE}$ (error in average treatment effect estimation) and $\epsilon_{PEHE}$ (error in precision for estimating heterogeneous effects) were used for evaluation. As illustrated in Table~\ref{table:Bayes}, CausalBGM demonstrated competitive performance in $ATE$ estimation, achieving the best performance in 3 out of 9 datasets. For example, CausaBGM achieves the lowest error of $0.0061$ in the first dataset with sample size $1k$, far surpassing the second best method CausalEGM by 37.1\%. However, CausalEGM remained the leading method for $ATE$ estimation on large datasets, such as those with sample sizes of 50k, indicating its robustness in handling extensive data. In contrast, CausalBGM demonstrated superior performance in estimating $\epsilon_{PEHE}$ that considers the individual treatment effects (ITEs). CausalBGM demonstrated superior performance by achieving the best results in 8 out of 9 datasets. The improvements were particularly substantial in specific datasets. For instance, CausalEGM reduced the error by 2.4 folds in the first dataset with sample size $1k$ comparing to the second best method. It achieved a 2.3 folds improvement in the last dataset with sample size $50k$. 

Overall, these results demonstrate the robustness, scalability, and precision of CausalBGM, particularly excelling in individual treatment effect estimation. Its substantial improvements over strong baselines underscore its potential as a state-of-the-art approach for causal inference tasks.

\renewcommand\arraystretch{0.7}
\begin{table}
  \centering
  \resizebox{\textwidth}{!}{
  \begin{tabular}{cccccccccc}
    \toprule
    Metric & Dataset & TARNET & CFRNET & CEVAE & GANITE & Dragonnet & CausalForest & CausalEGM & \textbf{CausalBGM} \\
    \midrule
    \multirow{9}{*}{$\epsilon_{ATE}$} & \multirow{3}{*}{Datasets-1k} 
    & $0.022 \pm 0.015$ & $0.018 \pm 0.015$ & $0.035 \pm 0.021$ & $0.27 \pm 0.08$ & $0.010 \pm 0.004$ & $0.021 \pm 0.001$ & $0.0097 \pm 0.0075$ & $\bm{0.0061\pm0.0041}$ \\
    & & $0.038 \pm 0.029$ & $0.041 \pm 0.027$ & $0.12 \pm 0.10$ & $2.0 \pm 0.3$ & $\bm{0.012 \pm 0.007}$ & $0.017 \pm 0.003$ & $0.032 \pm 0.020$ & $0.029 \pm 0.028$ \\
    & & $0.10 \pm 0.06$ & $\bm{0.095 \pm 0.079}$ & $0.38 \pm 0.27$ & $2.0 \pm 1.4$ & $0.16 \pm 0.10$ & $0.23 \pm 0.02$ & $0.26 \pm 0.07$ & $0.13\pm0.05$ \\
    \cline{2-10}
    & \multirow{3}{*}{Datasets-10k} 
    & $6.4 \pm 3.5$ & $12 \pm 7$ & $204 \pm 58$ & $2.7 \pm 1.2$ & $124 \pm 11$ & $2.5 \pm 1.1$ & ${1.3 \pm 0.6}$ & $\bm{1.22\pm0.80}$ \\
    & & $0.056 \pm 0.001$ & $0.056 \pm 0.001$ & $0.070 \pm 0.031$ & $1.2 \pm 0.2$ & $0.0097 \pm 0.069$ & $0.0057 \pm 0.0004$ & $0.0043 \pm 0.0025$ & $\bm{0.0038\pm0.0029}$ \\
    & & $0.034 \pm 0.023$ & $0.060 \pm 0.002$ & $0.018 \pm 0.011$ & $0.12 \pm 0.09$ & $0.078 \pm 0.057$ & $\bm{0.013 \pm 0.003}$ & $0.039 \pm 0.016$ & $0.019\pm0.018$ \\
    \cline{2-10}
    & \multirow{3}{*}{Datasets-50k} 
    & $0.038 \pm 0.021$ & $0.085 \pm 0.105$ & $0.59 \pm 0.31$ & $1.4 \pm 0.5$ & $0.89 \pm 0.53$ & $0.024 \pm 0.003$ & $\bm{0.020 \pm 0.013}$ & $0.045\pm0.015$ \\
    & & $0.044 \pm 0.003$ & $0.045 \pm 0.004$ & $0.66 \pm 0.59$ & $2.3 \pm 0.2$ & $0.027 \pm 0.028$ & $0.010 \pm 0.001$ & $\bm{0.0098 \pm 0.0089}$ & ${0.010\pm0.003}$ \\
    & & $0.30 \pm 0.01$ & $0.30 \pm 0.01$ & $0.64 \pm 0.45$ & $1.9 \pm 0.3$ & $0.16 \pm 0.08$ & $0.12 \pm 0.01$ & $\bm{0.0016 \pm 0.0010}$ & $0.012\pm0.009$ \\
    \midrule
    \multirow{9}{*}{$\epsilon_{PEHE}$} & \multirow{3}{*}{Datasets-1k} 
    & $0.11 \pm 0.02$ & ${0.00069 \pm 0.00075}$ & $0.012 \pm 0.005$ & $0.14 \pm 0.04$ & $0.038 \pm 0.003$ & $0.00080 \pm 0.00005$ & $0.0069 \pm 0.0016$ & $\bm{0.00029 \pm 0.00020}$ \\
    & & $0.35 \pm 0.03$ & $0.29 \pm 0.04$ & $0.27 \pm 0.04$ & $4.34 \pm 1.24$ & $0.34 \pm 0.01$ & $0.27 \pm 0.01$ & $0.25 \pm 0.01$ & $\bm{0.18\pm0.01}$ \\
    & & $0.31 \pm 0.14$ & $0.28 \pm 0.23$ & $7.6 \pm 5.3$ & $12 \pm 6$ & $1.7 \pm 0.4$ & $\bm{0.075 \pm 0.006}$ & $0.20 \pm 0.03$ & $0.12\pm0.03$ \\
    \cline{2-10}
    & \multirow{3}{*}{Datasets-10k} 
    & $433 \pm 106$ & $662 \pm 288$ & $46200 \pm 15500$ & $78.7 \pm 26.8$ & $22200 \pm 4130$ & $483.72 \pm 31.68$ & ${7.2 \pm 2.6}$ & $\bm{6.23\pm1.92}$ \\
    & & $0.024 \pm 0.005$ & $0.022 \pm 0.006$ & $0.091 \pm 0.019$ & $2.08 \pm 0.45$ & $0.042 \pm 0.003$ & $0.015 \pm 0.001$ & $0.014 \pm 0.001$ & $\bm{0.0120\pm0.0001}$ \\
    & & $0.012 \pm 0.005$ & $0.0040 \pm 0.0028$ & $0.0034 \pm 0.0013$ & $0.14 \pm 0.08$ & $0.036 \pm 0.015$ & $0.0016 \pm 0.0008$ & $0.0028 \pm 0.0013$ & $\bm{0.0010\pm0.0014}$ \\
    \cline{2-10}
    & \multirow{3}{*}{Datasets-50k} 
    & $0.88 \pm 0.04$ & $0.90 \pm 0.08$ & $1.1 \pm 0.5$ & $3.4 \pm 1.4$ & $1.84 \pm 0.83$ & $0.65 \pm 0.01$ & $0.55 \pm 0.01$ & $\bm{0.54\pm0.01}$ \\
    & & $0.031 \pm 0.006$ & $0.030 \pm 0.011$ & $0.84 \pm 0.76$ & $5.454 \pm 0.65$ & $0.039 \pm 0.007$ & $0.020 \pm 0.002$ & $0.022 \pm 0.001$ & $\bm{0.019\pm0.001}$ \\
    & & $0.22 \pm 0.07$ & $0.27 \pm 0.05$ & $0.67 \pm 0.61$ & $3.8 \pm 1.1$ & $0.14 \pm 0.06$ & $0.022 \pm 0.001$ & $0.0054 \pm 0.0013$ & $\bm{0.0024\pm0.0011}$ \\
    \bottomrule
  \end{tabular}}
 
\caption{Binary treatment experiments with CausalBGM and competing methods on the ACIC 2018 datasets with varying sample size. Each method was run 10 times, and the standard deviations are shown. The best performance is marked in bold.}
\label{table:Bayes}
\end{table}

Furthermore, we evaluate whether the CausalBGM framework can learn a more effective low-dimensional representation compared to CausalEGM and sufficient dimension reduction (SDR). It is important to note that all SDR-based methods for causal inference rely on linear SDR, which is inherently restrictive and may fail to capture nonlinear relationships in complex datasets. To assess this, we conducted a comprehensive comparison of CausalBGM with SDRcausal under experimental settings that either satisfied or violated the SDR assumption. SDRcausal implements several variants proposed in the original study \citep{sdrCausal2021}, and for fairness, we always report the best-performing result. CausalBGM demonstrated significant improvements over SDRcausal in both experimental settings, with substantial advantages in nonlinear datasets where the linear SDR approach was unable to model the underlying complexity effectively (see Appendix E). These results highlight the capability of CausalBGM to overcome the limitations of linear assumptions and better capture intricate relationships in high-dimensional data.

\subsection{Posterior Interval}
Unlike most of the existing methods that only focus on point estimation, CausalBGM adopts the Bayesian inference principle, thus enabling uncertainty quantification and providing posterior interval of the causal effect estimates. More importantly, since the latent features are inferred for each subject, CausalBGM is able to offer individual treatment effect estimate with a posterior interval. To assess the utility of the posterior interval, we evaluate it based on its coverage probability or empirical coverage, which involves checking how often the true causal effect (e.g., average dose-response) lies within the predicted interval from a frequentist perspective.

\begin{figure}[htbp]
  \centering
  \includegraphics[width=0.9\columnwidth]{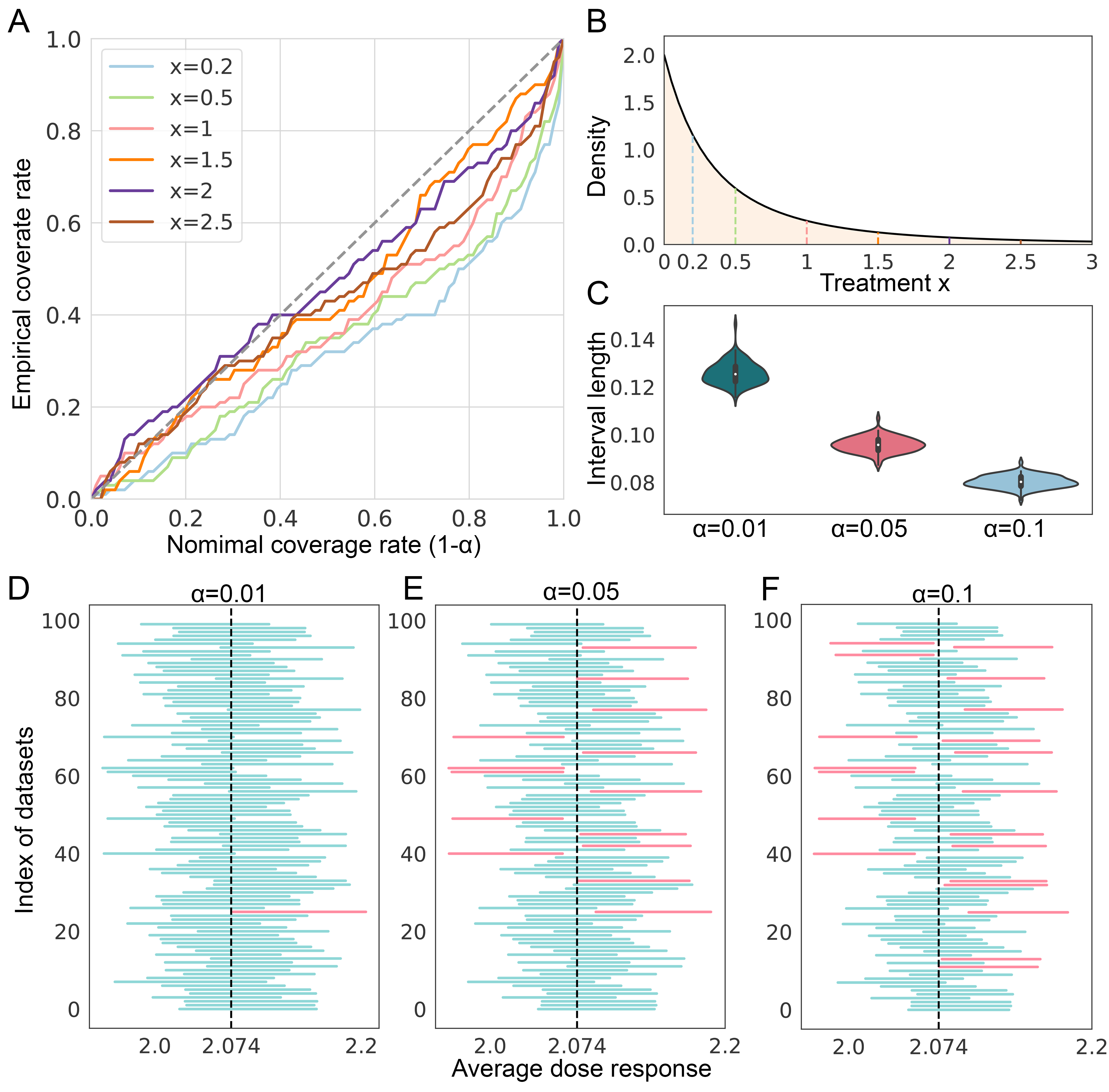}
  \caption{posterior interval analysis of CausalBGM using Imbens et al. dataset. (A) The calibration of empirical coverage rate at different treatment values ($x=0.2,0.5,1,1.5,2,2.5$). (B) The marginal density plot of treatment value $x$. Vertical dotted lines with different colors represent different treatment values ($x=0.2,0.5,1,1.5,2,2.5$) (C) The distribution of interval lengths at different significant levels $\alpha=0.01,0.05,0.1$. (D-F) The coverage indicator plots of CausalBGM at different significant levels $\alpha=0.01,0.05,0.1$ where the horizontal line indicates the truth average dose-response value at $x=2$, the “covered” intervals are marked in green, and “missed” intervals are marked in red.}
  \label{fig:prediction_interval}
\end{figure}

We used the Imbens et al. dataset as a case study to evaluate the empirical coverage rate of posterior intervals estimated by CausalBGM. Specifically, 100 independent datasets were generated using different random seeds, and CausalBGM was applied to each dataset to estimate the average dose-response function $\mu(x)$. For a given treatment value $x$, the empirical coverage rate was defined as the proportion of times a posterior interval successfully contains the true value at a specific significant level $\alpha$. By varying the significant level $\alpha$, we generated a calibration curve of the empirical coverage rate. Interestingly, the empirical coverage rate was more accurate at treatment values $x=1.5, 2$ compared to other treatment values (Figure~\ref{fig:prediction_interval}A). This discrepancy across different treatment values can be attributed to 1) the treatment value distribution as shown in the marginal density plot of $x$ (Figure~\ref{fig:prediction_interval}B). 2) The property (e.g., slope) of the truth average dose-response curve (Figure~\ref{fig:continuous_result}B). To further investigate, We took the $x=2$, the best-performing case, as a focused study. By setting the significance level $\alpha$ to 0.01, 0.05, 0.1, the average length of the posterior interval decreased from 0.126 to 0.096 and 0.080, respectively (Figure~\ref{fig:prediction_interval}C). Additionally, we visualized the 100 posterior intervals of the average dose-response at treatment value $x=2$. As expected, smaller significant level $\alpha$ corresponded to a higher empirical coverage (Figure~\ref{fig:prediction_interval}D-F). For example, at $\alpha=0.01$, only one out of 100 intervals failed to cover the truth average dose-response value, highlighting the robustness of CausalBGM in providing accurate and well-calibrated posterior intervals.

\subsection{Effect of Initialization}
\label{sec:init}

The EGM initialization strategy plays an important role in ensuring the superior performance of CausalBGM. To evaluate the contribution of the EGM initialization strategy, we conducted a series of experiments comparing it with the traditional Xavier uniform initializer \citep{pmlr-v9-glorot10a} across three simulation datasets and one semi-synthetic dataset under the continuous treatment setting. The results, summarized in Table~\ref{table:init}, demonstrate that EGM initialization significantly enhances the performance of CausalBGM in terms of both RMSE and MAPE. Quantitatively, EGM initialization consistently reduced RMSE across all datasets and improved MAPE in three out of four datasets. For instance, in the Lee et al. dataset, EGM initialization achieved remarkable reductions in RMSE and MAPE by 93.4\% and 80.1\%, respectively. Similarly, in the Imbens et al. and Sun et al. datasets, EGM initialization substantially improved performance, with RMSE reductions of 70.5\% and 78.4\%, respectively. Even in the Twins dataset, where the impact on MAPE was marginal, EGM initialization still achieved a noticeable RMSE improvement of 50.0\%.

These findings underscore the critical importance of proper initialization strategies in enhancing the predictive accuracy and stability of CausalBGM. By initializing the model parameters using the EGM strategy, CausalBGM effectively improved the prediction performance. Given the consistent improvements across multiple datasets, we adopt EGM initialization as the default strategy for the CausalBGM framework.

\renewcommand\arraystretch{0.6}
\begin{table}
 \caption{Effect of initialization strategy on the performance of CausalBGM. Note that CausalBGM adopts EGM initialization strategy by default. CausalBGM$\ast$ represents CausalBGM without EGM initialization and directly adopts the Xavier uniform initializer. Each method was run 10 times and the standard deviations are shown}
  \centering
  \begin{tabular}{ccccl}
    \toprule
    Dataset&Method&RMSE&MAPE\\
    \midrule
    \multirow{2}*{Imbens et al.}
    &CausalBGM$\ast$&$0.095\pm0.009$&$0.025\pm0.006$\\
    &CausalBGM&$\bm{0.028\pm0.007}$&$\bm{0.013\pm0.003}$\\
    \midrule
    \multirow{2}*{Sun et al.}     
    &CausalBGM$\ast$&$0.171\pm0.080$&$0.054\pm0.012$\\
    &CausalBGM&$\bm{0.037\pm0.009}$&$\bm{0.013\pm0.005}$\\
    \midrule
    \multirow{2}*{Lee et al.}
    &CausalBGM$\ast$&$1.221\pm0.128$&$0.362\pm0.017$\\
     &CausalBGM&$\bm{0.080\pm0.030}$&$\bm{0.072\pm0.035}$\\
    \midrule
    \multirow{2}*{Twins} 
    &CausalBGM$\ast$ &$0.062\pm0.018$&$\bm{0.067\pm0.024}$\\
     &CausalBGM&$\bm{0.031\pm0.007}$&${0.077\pm0.009}$\\
  \bottomrule
\end{tabular}
\label{table:init}
\end{table}

\subsection{Scalability}

Scalability has become a critical requirement in causal inference, particularly for modern applications involving increasingly large and complex datasets. To evaluate the scalability of CausalBGM, we conducted comprehensive experiments examining its ability to handle datasets with (i) increasing the number of covariates and (ii) increasing the sample size (see Appendix~F). We first fixed number of covariates and increased the sample size using Imbens et al dataset. Only OLS, CausalEGM and CausalBGM can scale to million-sample dataset (e.g., $N=10^6$) while other approaches failed due to memory or time limitations. We then fixed the sample size and increased the number of covariantes from $p=50$ to $p=10{,}000$. In this setting, several competing methods (e.g., DML) failed to run as $p$ increased while CausalBGM demonstrated consistently strong performance across different range of covariate dimensions. We further showed the computational time of CausalBGM by varying sample size over orders of magnitude from $N=10^3$ to $10^6$ (See Appendix G). 

\section{Discussion}
\label{sec:dis}

In this article, we introduced CausalBGM, an AI-powered Bayesian generative modeling framework for causal inference, particularly excelling in observational studies with high-dimensional covariates and large-scale datasets. By combining deep generative modeling with Bayesian principles and causal reasoning, CausalBGM provides a flexible and robust approach for flexible covariate adjustment, individualized effect estimation, and principled uncertainty quantification.

One notable contribution of CausalBGM is the ability to estimate individual-specific posterior distributions of latent features (interpretable as latent confounders), enabling posterior intervals for individual treatment effects (ITEs) in addition to point estimates. This individualized posterior representation offers a practically useful and statistically grounded way to quantify uncertainty at the individual level, which remains underdeveloped in existing causal inference methods. Additionally, CausalBGM is designed to be scalable since the iterative updating algorithm only requires a mini-batch of samples at each step. Such scalability, combined with its robust statistical foundations, makes CausalBGM a practical and powerful tool for addressing the demands of modern applications in genomics, healthcare, and social sciences.

Despite these strengths, several limitations and open questions remain, and we explicitly address key methodological claims to improve transparency, reproducibility, and interpretability. First, The latent variables are typically unidentifiable for latent variable modeling (LVM) since different latent representations can induce the same marginal distribution over the observed variables. We discuss this issue by leveraging recent advances in nonlinear independent component analysis (ICA) theory (see Appendix H). Specifically, identifiability can be achieved under mild conditions by introducing an auxiliary variable and conditioning the latent prior on it. Building on this idea, we developed an identifiable variant, iCausalBGM and evaluated the empirical performance compared to CausalBGM.

Second, using variational inference (VI) may lead to an underestimation of the posterior uncertainty \citep{murphy2012machine}, we implement a ``full MCMC'' version of CausalBGM where Metropolis–Hastings is used for sampling individual latent variables and Hamiltonian Monte Carlo is used for sampling model parameters. The empirical performance is benchmarked against the original CausalBGM (see Appendix I). 

Third, CausalBGM relies on a latent-variable version of unconfoundedness, which requires relevant confounding information can be captured by observed covariates. When important covariates are missing, the impact depends on their causal role. We evaluate the performance impact on CausalBGM under different ``missing-covariate'' scenarios to demonstrate the robustness of our method (see Appendix J). 

Looking forward, several directions could further strengthen the framework. On the theory side, while CausalBGM demonstrates strong empirical performance with its Bayesian foundation, further theoretical work is needed to rigorously characterize the convergence properties of CausalBGM under the proposed iterative algorithm. On the methodological side, the sensitivity of CausalBGM to parameter initialization remains poorly understood, which could limit its adaptability in scenarios where the EGM framework is less effective. Investigating the underlying causes of this sensitivity and exploring alternative initialization strategies or adaptive learning mechanisms could further enhance the robustness and versatility of the framework. 

In conclusion, CausalBGM provides a new perspective on developing Bayesian causal inference methods by harnessing the power of AI. Its flexibility, scalability, and strong empirical performance make it a valuable tool for a wide range of applications in modern data science. 

\section{Funding}
Q.L. was partially supported by NIH grant R00 HG013661. W.H.W. was partially supported by NSF-DMS 2310788. 

\section{Disclosure Statement}
The authors report there are no competing interests to declare.









\bibliographystyle{chicago.bst}

\bibliography{Bibliography-MM-MC}
\end{document}